\documentclass{article}

\usepackage{microtype}
\usepackage{graphicx}
\usepackage{subfigure}
\usepackage{booktabs}     
\usepackage{array}        
\usepackage{tabularx}     
\usepackage{siunitx}      
\usepackage[breaklinks=true]{hyperref}

\usepackage[accepted]{icml2025}

\usepackage{amsmath}
\usepackage{amssymb}
\usepackage{mathtools}
\usepackage{amsthm}

\usepackage[capitalize,noabbrev]{cleveref}

\theoremstyle{plain}

\theoremstyle{definition}

\theoremstyle{remark}

\usepackage[textsize=tiny]{todonotes}

\icmltitlerunning{Attestable Audits: Verifiable AI Safety Benchmarks Using Trusted Execution Environments}

\DeclareRobustCommand*\circled[1]{\tikz[baseline=(char.base)]{
            \node[shape=circle,fill=black,fill,inner sep=1.2pt] (char) {\scriptsize \textcolor{white}{\textbf{\sffamily{#1}}}}}}

\begin{document}

\twocolumn[
\icmltitle{Attestable Audits: Verifiable AI Safety Benchmarks \\ Using Trusted Execution Environments} 



\icmlsetsymbol{equal}{*}

\begin{icmlauthorlist}
\icmlauthor{Christoph Schnabl}{cam}
\icmlauthor{Daniel Hugenroth}{cam}
\icmlauthor{Bill Marino}{cam}
\icmlauthor{Alastair R. Beresford}{cam}
\end{icmlauthorlist}

\icmlaffiliation{cam}{Department of Computer Science and Technology, University of Cambridge}

\icmlcorrespondingauthor{Christoph Schnabl}{cs2280@cst.cam.ac.uk}

\icmlkeywords{Machine Learning, ICML}
\vskip 0.3in
]



\printAffiliationsAndNotice{}  

\begin{abstract}
Benchmarks are important measures to evaluate safety and compliance of AI models at scale. However, they typically do not offer verifiable results and lack confidentiality for model IP and benchmark datasets. We propose Attestable Audits, which run inside Trusted Execution Environments and enable users to verify interaction with a compliant AI model. Our work protects sensitive data even when model provider and auditor do not trust each other. This addresses verification challenges raised in recent AI governance frameworks. We build a prototype demonstrating feasibility on typical audit benchmarks against Llama-3.1.
\end{abstract}

\section{Introduction}
\label{sec:introduction}
Audits are an essential tool in the modern AI safety landscape as models become more capable~\cite{aschenbrenner2024situational} and potentially more dangerous~\cite{barrett2023actionable, anthropic2023a, openai2024a}, particularly in agentic environments~\cite{harms2023agentic}. Recognizing these risks, several AI regulations~\cite{EU2024, coloradoga2024, whitehouse2023a}, policy initiatives, and AI principles~\cite{houseofcommons2024ai, solaiman2023release, kapoor2024societal} have mandated audits, but decision-makers often lack the technical information needed to evaluate auditing tools~\cite{reuel2024open}. This creates a critical gap between policy and implementation that Technical AI Governance aims to close through tools, such as verifiable audits~\cite{reuel2024open}. However, current audits rely on contracts or manual processes, and verification remains challenging due to restricted model access and data privacy concerns~\cite{longpre2024safeharbor, south2024, carlini2024stealing, cen}. Furthermore, misaligned incentives between stakeholders can result in audits that do not serve the public interest~\cite{casper2024audit, raji2022, mokander2023}, including data exfiltration by involved actors~\cite{eriksson2025trustaibenchmarksinterdisciplinary} or models that intentionally underperform during evaluations~\cite{vanderweij2025aisandbagginglanguagemodels}.

To address these challenges, we investigate how users can verify they are interacting with a compliant AI system under realistic constraints: when model providers do not share weights, auditors only share code and data with regulators, and systems run on untrusted third-party infrastructure. 

We propose Attestable Audits (\S\ref{sec:attested-audits}), a three-step verification protocol, where auditors and model providers securely load models, audit code, and datasets into hardware-backed Trusted Execution Environments (TEEs), run benchmarks, and cryptographically attest and publish results to a public registry for user verification. We use TEEs from Confidential Computing (CC, \S\ref{sec:confidential}) to isolate execution and encrypt memory. We demonstrate through a prototype (\S\ref{sec:evaluation}) based on AWS Nitro Enclaves that benchmarks yield expected results at 2.2$\times$ the cost of CPU and 21.7$\times$ that of GPU inference.


\section{Confidential Computing (CC)}
\label{sec:confidential}

\begin{figure*}[th]
\vskip 0.2in
\begin{center}
\centerline{\includegraphics[width=\textwidth]{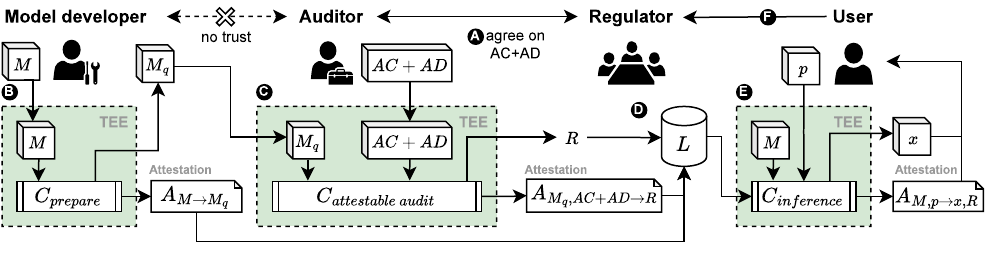}}

\caption{
Overview of the Attestable Audit protocol. 
\circled{A} The auditor and regulator agree on audit code $AC$ and dataset $AD$.
\circled{B}~Optionally, the provider prepares a (quantized) version $M_q$ of model $M$, verifiable via attestation $A_{M \rightarrow M_q}$.
\circled{C} In the audit, the auditor loads encrypted $AC+AD$ into a fresh TEE and the provider loads encrypted $M$.
\circled{D} The audit result $R$ and attestation $A_{Mq,AC+AD \rightarrow R}$ are published to a transparency log $L$.
\circled{E} The user confidentially sends prompt $p$ to $M$ (or $M_q$) in a TEE, receiving $x$ and an attestation verifying provenance and $M$'s compliance.
\circled{F} The user may disclose $(p,x,A_{M,p \rightarrow x,R})$ to the regulator to show audit deficits.
}
\label{fig:flow}
\end{center}
\vskip -0.2in
\end{figure*}

Confidential Computing (CC) ensures that critical systems protect data-in-transit, data-at-rest, and data-in-use.  
This is achieved by TEEs, privileged execution modes supported by modern CPUs---conceptually, a small, shielded, encrypted computer inside a computer.  
Backed by secure hardware, TEEs prevent interference by the host/hypervisor and encrypt all memory to thwart even physical attacks by attackers.  
This makes CC attractive for deployments at otherwise untrusted cloud service providers (CSPs) as long as the vendor of the secure hardware is trusted~\cite{chen2023verified}.

In contrast to first generation process-based TEEs (e.g., Intel SGX, Arm TrustZone), second generation TEEs (e.g., AMD SEV-SNP, Intel TDX, AWS Nitro) support full VMs~\cite{costan2016intelsgxexplained,pinto2019armtrustzoneeplained,inteltdx,amdsevsnp}.
This helps them overcome resource limitations, especially memory, that previously made ML workloads difficult~\cite{mo2024machine}.
Whereas many CC applications focus on confidentiality properties, our work also leverages its integrity guarantees which can provide zero-knowledge-proof-like guarantees~\cite{russinovich2024confidential}.

Clients can verify they are talking to a service inside a TEE through Remote Attestation (RA).
In RA, the secure chip signs a chain of measurements, called Platform Configuration Registers (PCRs), using a non-extractable secret key.
The PCRs cover the Trusted Computing Base (TCB), consisting of firmware and the loaded enclave base image. Including the enclave base image enables revocation when vulnerabilities, e.g., side-channels~\cite{li2021cipherleaks}, active attacks~\cite{schluter2024heckler}, and memory aliasing~\cite{badramsp25}, are discovered.

We use AWS Nitro Enclaves~\cite{aws_nitro_enclaves} as the CC platform for our prototype.
Our protocol is compatible with other CC platforms, such as Intel TDX and AMD SEV-SNP, but we leave those implementations for future work. 
We expect that alternatives allow for smaller TCBs and lower overhead. 
Importantly, these can also integrate with GPUs featuring CC support, such as Nvidia's H100~\cite{nvidia2023blog} to allow for larger models.
However, these eventually face limits, e.g., there is no multi-GPU support.
As such, our conservative choice of a smallest common-denominator technology, ensures our design supports these challenges.

\begin{table*}[ht]
  \vskip 0.15in
  \begin{center}
  \begin{small}
  \caption{Overview of common AI safety benchmarks. Underlined benchmarks were chosen as representative $AC+AD$ for our evaluation. }
  \label{tab:benchmarks}
  \begin{tabular}{@{}p{0.142\textwidth}@{\hspace{0.01\textwidth}}p{0.848\textwidth}@{}}
    \toprule
    \textbf{Type} & \textbf{Benchmarks} \\
    \midrule
    Discrete-Label
      & \underline{MMLU}~\cite{mmlu}, BoolQ~\cite{boolq}, HellaSwag~\cite{hellaswag} \\
    Text-Similarity
      & \underline{XSum}~\cite{xsum}, NarrativeQA~\cite{narrativeqa}, CNN/DailyMail~\cite{cnndailymail} \\
    Classifier-Judged
      & \underline{ToxicChat}~\cite{toxicchat}, BBQ~\cite{bbq}, RealToxicityPrompts~\cite{realtoxicityprompts}\\
      Retrieval
      & MSMARCO~\cite{msmarco}, Natural Questions~\cite{naturalquestions} \\
    LLM-as-a-Judge
    & Chatbot Arena~\cite{chatbotarena} \\
    \bottomrule
  \end{tabular}
  \end{small}
  \end{center}
  \vskip -0.1in
\end{table*}

\section{Attestable Audits}
\label{sec:attested-audits}

We present Attestable Audits as a three-step design depicted in Figure~\ref{fig:flow}. First, model providers may prepare their model. Then, providers and auditors load the audit code and data into a TEE, which runs benchmarks and publishes attested results. Finally, users receive output attestations to verify interaction with an audited model.

\subsection{Security Goals and Threat Model}
The efficacy of the proposed system relies on the following security goals. We require \textbf{model verifiability (G1)}, i.e., the attestation includes hashes of model weights and code, and \textbf{audit verifiability (G2)}, i.e., outputs are bound to an approved audit version. The system must maintain \textbf{confidentiality (G3)} of model weights (protecting IP) and audit data to prevent ``cheating''. These guarantees are hard to achieve in non-CC setups but enable more robust audits, especially for closed-source systems. \textbf{Transparency (G4)} requires publishing base image, model, and audit digests, with verifiable build steps. Finally, the system must enforce \textbf{statelessness (G5)} to prevent prompt residue and covert channels~\cite{shumailov2025tcme}, and \textbf{output verifiability (G6)} to authenticate model responses during interaction.

We assume the existence of \textbf{Network adversaries (A1)} who can intercept, tamper with, or spoof communication between components, but exclude DoS attacks. For \textbf{Physical and Privileged adversaries (A2)}, with capabilities such as RAM snapshots, VM rollbacks, and side-channel attacks. 

\subsection{Requirements}

We use three standard cryptographic primitives available in libraries like \textsc{LibSodium}. First, we require a pre-image and collision-resistant hashing function. Second, we require an IND-CCA secure key encapsulation mechanism (KEM) to allow two parties to share a symmetric key. The receiver generates $(pk, sk)  \gets \textsc{KEM.KeyGen()}$ and shares $pk$. Then sender uses $pk$ to generate a key and ciphertext $c, k \gets \textsc{KEM.Encapsulate}(pk)$. The receiver recovers $k$ using $\textsc{KEM.Decapsulate}(sk, c)$. Thirdly, we require an IND-CCA secure encryption scheme (AEAD) with $c_x \gets \textsc{AEAD.encrypt}(k, x)$ to encrypt plaintext $x$ under key $k$, and $x \gets \textsc{AEAD.decrypt}(k, c_x)$ to decrypt.

Furthermore, we rely on functionality provided by the TEE. First, we require attestation $(\{d, \dots\}, \textsc{pcr}, \sigma) \gets \textsc{Attest}(\{d, \dots\})$ against the currently running TEE image.
The attestation includes (1) the platform configuration registers $\textsc{pcr}$ that describe the loaded image, (2) auxiliary user-provided data $\{d, \dots\}$, and (3) a signature $\sigma$ over all these signed with the TEE vendor's secret key. We denote attestations as \( A_{in \rightarrow out} \), for binding hashed input \( in = \textsc{Hash}(input) \) to hashed output \( out = \textsc{Hash}(output) \).

We run the model code in a sandbox for isolation. Public model code can be included in the attested open-source base image, and only the weights need to remain confidential.

\section{Protocols}
\label{sec:protocols}

This section describes the main protocols of our Attestable Audit scheme.
Appendix~\ref{sec:appendix:protocol-requirements} describes used primitives and contains the pseudocode listings.
The shown protocols omit details, e.g. replay prevention and key rotation mechanisms, that are important for real-world implementations.

\textbf{The \textsc{Prepare} protocol} (Algorithm~\ref{alg:prepare}) offers model developers the ability to quantize their models in a confidential and verifiable manner. Practically, it enables the use of substantially smaller models with comparable performance. We discuss this ablation in Appendix~\ref{sec:appendix:quantize}. Conceptually, it illustrates the general, abstract attestation-and-encryption workflow, based on \textsc{KEM} and \textsc{AEAD}, that we likewise employ in our subsequent protocols.

First the TEE boots from its secure image and generates a fresh KEM keypair $pk, sk$.
It then attests to its boot image and the public key with a fresh attestation $A$ to allow third parties to verify the given $pk$ was indeed generated inside a TEE that booted a trusted image. The third-party first compares the $\textsc{pcr}$ measurement against known trusted images and then verifies the signature $\sigma$ using the TEE's vendor public key (or respective attestation service).

Once the authenticity of the TEE has been confirmed, the model provider calls $(k, c) \gets \textsc{KEM.Encapsulate}(pk)$ and uses $k$ to encrypt their model $M$.
The encrypted model $c_M$ and the encrypted key $c$ are then sent to the TEE which can derive the same key using $sk$ and decrypt the model.
The TEE then computes the quantized model $M_q$ and measures boths by computing their hashes $h_M$ and $h_{M_q}$.

Finally, the TEE encrypts the quantized model $M_q$ using the same symmetric key $k$ and sends it to the model provider.
It also publishes the attestation $A_{M\rightarrow M_q}$ to a transparency log. This acts as evidence for third parties that they can accept models with the hash $h_{M_q}$ as quantized versions of models with the hash $h_M$.
The $\textsc{Prepare}$ step is convenient for practical deployments as some TEEs, e.g. the one for the audit, can only run against smaller models $M_q$. It allows the model provider to deploy the full model $M$ while convincing others that the audit of $M_q$ is a valid approximation.

\textbf{The $\textsc{AttestableAudit}$ protocol} (Algorithm~\ref{alg:attestable-audit}) runs audit code $AC$ and audit dataset $AD$ against a (quantized) model $M_q$ in a confidential and verifiable manner.
For this the TEE, as in the previous protocol, publishes an attestation with a KEM public key.
Now both the model developer and the audit provider use it to upload the encrypted model $c_{M_q}$ and audit code/dataset $c_{AC+AD}$ to the TEE.

Once the TEE has received both, it will create a sandbox and executes $AC$ against $M_q$ using the audit dataset $AD$.
The sandbox ensures that malicious code that is part of $AC$ or $M_q$ cannot interfere with the integrity of the overall TEE logic or invalidate previous measurements.
After the execution, $AC$ will output a single aggregated result $R$.
The TEE then publishes an attestation $A_{M_q,AC,AD\rightarrow R}$ that binds the hash of the model and audit to this result $R$.
This is then published to a transparency log.

\textbf{The $\textsc{Inference}$ protocol} (Algorithm~\ref{alg:inference}) allows users to confidentially interact with model $M$ (or $M_q$) confidentially while being able to receive guarantees that it has received score $R$ against the audit $AC+AD$.
Different to the previous protocols, the TEE first downloads the relevant attestation documents from the previous steps from the transparency logs and includes these in its initial attestation.
That provides a baseline guarantee to the user that their later prompts are not given to a different model.

Next, the model provider loads their model $M$ into the TEE using the familiar KEM+AEAD construction.
The TEE will verify that the hash of the model matches with the value from the attestations and abort if that is not the case.

Then the user sends the encrypted prompt $c_p$ to the TEE.
The TEE then starts a sandbox with the model $M$ and runs it against the input $p$ yielding output $x$.
A fresh attestation $A_{M,p \rightarrow x,R}$ then provides the user also with a evidence that links together model, audit result, prompt, and response.
Optionally the user can later use this attestation to proof short-comings in the auditing process to a regulator.

\section{Evaluation}
\label{sec:evaluation}

We sample three AI‐safety benchmarks (Table~\ref{tab:benchmarks}) to assess Attestable Audits' feasibility. With additional engineering, we could integrate larger benchmark suites, such as COMPL-AI~\cite{complai} and HELM~\cite{helm}. We already employ zero-shot prompts from both. Our implementation, written in Rust through bindings to \texttt{llama.cpp}~\cite{llamacpp}, runs on AWS Nitro Enclaves using Llama-3.1-8B-Instruct quantized to 4-bit to reduce main memory footprint at an accuracy penalty.

We log input/output token counts and prompt/response decoding latencies, ignoring non-decoding overhead (e.g., copying models into the enclave takes \(\le\)2 minutes). We issue 500 prompts per benchmark on: (I) a \texttt{m5.2xlarge} instance running our protocol with enclaves enabled on 4 cores, (II) a version on \texttt{m5.xlarge} running on 4 cores, (III) a cost-constant version running \texttt{m5.2xlarge} on 8 cores, and (IV) a SOTA baseline hosting the same model, but in \texttt{fp16} precision on a cloud-based NVIDIA L40S, which uses roughly 90\% of the available 48 GB VRAM through vLLM at a price of 0.89 USD/hour.
\paragraph{Feasibility}
\label{sec:feasibility}
We demonstrate that models in Attestable Audits achieve adequate performance (Table~\ref{tab:combined-table}). In column (I), the quantized model’s zero-shot MMLU accuracy is 51.4\% (57.4\% excluding unparsable responses), similar to 54.6\% on a non-quantized model (IV) at 58.9\% and LLaMa’s 66.7\% for 5-shot prompting~\cite{touvron2023llama}. The difference from (I--III) to (IV) is context size and precision; from (IV) to LLaMa’s 66.7\% is prompting. Summarization yields a mean BERT score of $\approx$0.47 vs.\ $\approx$0.58 for the non-quantized version. On ToxicChat, 1.78\% are jailbreak attempts. The quantized model fails to refuse and produces toxic outputs in 2.4\% of all test cases in (I) and 2.6\% in the non-quantized case. Smaller differences between (I--III) stem from stochastic \texttt{top\_p} sampling. A 4-bit quantized model slightly degrades performance in benchmarks. We provide a more detailed feasibility analysis in Appendix~\ref{sec:benchmark-feasibility}.

\begin{table}[ht]
  \caption{Trade-offs and benchmark scores across: (I) enclave, (II) compute-constant, (III) cost-constant, vs. (IV) L40S GPU}
  \label{tab:combined-table}
  \vskip 0.15in
  \begin{center}
  \begin{small}
  \begin{sc}
    \sisetup{
      table-space-text-post = {\%},
      round-mode = places,
      round-precision = 2
    }
    \begin{tabular}{
      p{3.4cm} @{
      \hskip 8pt} 
      S[table-format=2.2] @{
      \hskip 8pt} 
      S[table-format=2.2] @{
      \hskip 8pt} 
      S[table-format=2.2] @{
      \hskip 8pt}
      S[table-format=3.2]
    }
    \toprule
    Metric                   & {(i)}   & {(ii)} & {(iii)}  & {(iv)} \\
    \midrule
    Price/hr (\$)            & 0.38    & 0.19   & 0.38     & 0.89    \\
    Price/100k token (\$)    & 5.80    & 2.61   & 3.01     & 0.12    \\
    Token/s                  & 1.84    & 2.04   & 3.54     & 202     \\
    \midrule\midrule
    BERT Score               & 0.47    & 0.50   & 0.49     & 0.58    \\
    Toxicity rate (\%)       & 2.4     & 2      & 1.7      & 2.6     \\
    Accuracy (\%)            & 51.4    & 52.6   & 48.6     & 58.9    \\
    \bottomrule
    \end{tabular}
  \end{sc}
  \end{small}
  \end{center}
  \vskip -0.1in
\end{table}

\paragraph{Trade-Offs}
\label{sec:tradeoffs}

Running inference on CPUs incurs a cost overhead of 21.7$\times$ over GPU inference and suffers a 100$\times$ slowdown. The use of enclaves costs $2\times$ due to having to use a larger instance compared to (II) or sacrificing cores relative to (III). Memory capacity constraints forces the use of smaller or quantized models. (III) hosts a 4-bit model for comparability, but could host a 8-bit model for higher performance. Doubling the number of CPUs from (I) to (III) increases the throughput by almost 2\,token/s. We expect larger instances to reduce the overhead by 2–5$\times$.

\paragraph{Security}
\label{sec:security}

Model and audit verifiability (\textbf{G1, G2}) are achieved through the CC-powered \emph{audit step} (Algorithm~\ref{alg:attestable-audit}). 
The RA process binds the hashes of the quantised model \(M_q\), audit code \(AC\), and dataset \(AD\) with the platform PCRs measurements. Confidentiality (\textbf{G3}) is end-to-end through the use of ephemeral keys inside the TEE. The AEAD channel bound to the initial attestation protects data-in-transit from a network adversary (\textbf{A1}). VM-level enclave isolation and full-memory encryption deny physical attackers (\textbf{A2}) access to data-in-use. Transparency (\textbf{G4}) follows from publishing the enclave base image, build scripts, content hashes, and the attestations \(A_{M\rightarrow M_q}\), and \(A_{M_q,\,AC+AD\rightarrow R}\) to \(L\). Similar to Apple's PCC~\cite{apple2025pcc}, anyone can rebuild and inspect the exact evaluation environment. For statelessness (\textbf{G5}) each user session runs in a fresh VM-enclave that starts with zeroized RAM and not persistent storage to eliminate prompt residue. Output integrity (\textbf{G6}) is guaranteed in the \emph{interaction step} (Algorithm~\ref{alg:inference}) analogously. For each prompt \(p\) the enclave returns \((x,\,A_{M,\,p\rightarrow x,\,R})\), to bind \(\textsc{Hash}(M)\,\|\,p\,\|\,x\,\|\,R\).  Verifying \(A_{M,\,p\rightarrow x,\,R}\) against \(L\) confirms the reply is from the audited model with score \(R\). Prompt-based model exfiltration during the user interaction step remains a residual gap~\cite{carlini2024stealing}.

\paragraph{Engineering Challenges}
\label{sec:engineering}

Our approach works well for smaller (in terms of tokens) hand-crafted datasets, as CPU inference limits token throughput. Memory constraints of CC technology complicates hosting larger models. Enclaves do not have persistent memory but instead rely on a memory-mapped file system. As a result, the runtime memory requirements can be a multiple of the underlying base image. Due to this expansion, large files such as model weights or datasets have to be transferred into the enclave during runtime. Cryptographic operations, e.g., when establishing encrypted channels, add complex logic as well as additional CPU demands. Many CC platforms come with limited documentation and lack easy-to-use software libraries.
Additionally, a constant portion of main memory needs to be reserved for the host system and thus remains unavailable to the enclave, which is at least 64 MiB~\cite{aws_nitro_enclaves}, but for (I) is closer to 500 MiB, or 1.5\% of available memory.
\section{Related Work}

\paragraph{Hardware-Attested Integrity}
DeepAttest~\cite{chen2019deepattest} binds models and code to TEEs for CNNs but lacks audit traceability and targets on-premise. \citet{nevo2024securing} protect weights, while we extend this to audit datasets and code. \citet{openmined2024securellm} uses TEEs for evaluation but lacks generality and reproducibility. Our system integrates attestation into a transparent, regulator-facing pipeline.

\paragraph{Cryptographic Private Inference}
zkML~\cite{rath2024zkml} generates SNARKs for inference verification, but is orders of magnitude slower. FHE-LoRA~\cite{zhang2024fheLoRA} uses encrypted low-rank adaptation. SONNI~\cite{kim2025sonni} and Proof-of-Training~\cite{wang2023pot} focus on weight and data lineage, while we verify runtime behavior. PPFL~\cite{mo2021ppflprivacypreservingfederatedlearning} protects training gradients with TEEs, while we apply them to secure model evaluation.

\paragraph{Audit Frameworks \& Governance}
Audit Cards~\cite{auditcards2025} show audit gaps, especially in reproducibility and verification. \citet{mokander2024threeaudit} propose a layered taxonomy without cryptographic guarantees. \citet{grollier2025fairwashing} show audits can enable fairwashing. 
\citet{dong2024generalizationmemorizationdatacontamination} and \citet{leslie2023aifairness} focus on post-hoc safety, whereas we provide pre-deployment guarantees. \citet{brundage2020trustworthyaidevelopmentmechanisms} stress the importance of verifiable claims.

\section{Discussion \& Limitations}
\label{sec:discussion}

Our prototype has a high overhead in terms of runtime and costs.
Large parts of this can be attributed to the CPU-based inference that is required by the underlying CC technology.
A production-ready implementation using CC-compatible GPUs likely has an overhead as small as $5\times$.
We leave this, and the other suggested extensions below, for future work.

Our architecture requires the participating parties to trust the hardware vendor of the CC technology, in our case AWS.
However, all steps can be run on multiple CC technologies independently such that parties can later choose which attestation they trust.
We note that while this increases integrity guarantees (trust any), the model and audit confidentiality is reduced, as a single broken TEE can leak the sensitive data.

Our prototype requires a quantized model due to technical limitations of the chosen CC technology.
An implementation using a CC-compatible GPU can run the native Llama-3 model securely.
However, a preparation step will still be necessary for larger models that do not fit on a single GPU and can accommodate other post-training steps, e.g., fine-tuning.

While we presented our work in the context of LLMs, it generalizes to other AI/ML systems.
For instance, the operator of a self-driving car can use Attestable Audits to prove later, e.g., in court after an accident, that the very model that drove the vehicle at a given time has been correctly audited.
Note, that our infrastructure also prevents human-mistakes such as mixing up audit results or accidentally deploying a wrong model version.
It also naturally provides reproducibility for benchmarks during the scientific publication process.
\section{Conclusion}
\label{sec:conclusion}

Our Attestable Audits design uses TEEs to load audit code ($AC$), audit data ($AD$), and model weights ($M$) into an enclave, execute AI-safety benchmarks, and publish cryptographic proofs binding results to exact $AC$ + $AD$ + $M$ hashes. Our AWS Nitro Enclaves prototype runs three standard benchmarks at 21.7$\times$ the cost of GPU inference, with a CPU-constant variant at 2$\times$ overhead. Our protocol can be used for Verifiable Audits \cite{reuel2024open}~(\S5.4.1) without exposing sensitive IP. Our main limitation stems from CPU inference, but we expect this overhead to reduce as GPU-capable enclaves become more readily available. By shifting AI governance from ex post enforcement to ex ante certifiable deployment, Attestable Audits reduces compliance and transaction costs of governing AI systems.
\section*{Policy Brief} 
This work directly addresses a topic that is critical to the enforcement of any AI regulation, or, more broadly, any AI governance policy: when regulators or auditors lack direct access to a model or dataset due to privacy or competition concerns, how can they ensure it complies with the relevant requirements? By introducing a method for verifiable benchmarking of AI systems running on third-party infrastructure, we help bridge this gap and enable AI developers to provide clear assurances about their models and datasets to regulators, auditors, or any other stakeholders --- without exposing private assets such as model weights or proprietary data. By removing these barriers, our approach could incentivize more AI providers to enter regulated markets  \citep{reuters2023openai} or, differently, to enroll in voluntary pre-deployment testing by third parties \cite{field2024openai}.  Separately, because our method can help shift the cost of benchmark-based audits from resource-constrained \cite{aitken2022common} regulators to the AI developers who may be best-positioned to bear the expense of TEEs and, thus, the audits themselves, we lower the overhead associated with creating and enforcing AI regulations or AI governance policies. Given these benefits, future AI legislation might consider explicitly endorsing or encouraging such techniques.

\section*{Acknowledgements} 

We would like to thank the anonymous reviewers for their valuable feedback and suggestions, which helped improve this paper. Christoph is supported by Corpus Christi College, the German Academic Exchange Service, and the Studienstiftung des Deutschen Volkes. Daniel is supported by Nokia Bell Labs and Light Squares.

\section*{Impact Statement} 
Attestable Audits advance technical AI governance by allowing auditors, and users to securely verify AI model compliance. It mitigates risks of misuse or harmful outputs from AI systems through the use of trusted hardware that allows for rigorous audits without compromising sensitive data or proprietary models. While this enhances transparency and accountability, there remains a dependency on trusted hardware providers. Overall, we believe that Attestable Audits can reduce the risks associated with deploying powerful AI systems and contributes positively towards safer and more trustworthy machine learning applications.


\begin{thebibliography}{68}
\providecommand{\natexlab}[1]{#1}
\providecommand{\url}[1]{\texttt{#1}}
\expandafter\ifx\csname urlstyle\endcsname\relax
  \providecommand{\doi}[1]{doi: #1}\else
  \providecommand{\doi}{doi: \begingroup \urlstyle{rm}\Url}\fi

\bibitem[Aitken et~al.(2022)Aitken, Leslie, Ostmann, Pratt, Margetts, and
  Dorobantu]{aitken2022common}
Aitken, M., Leslie, D., Ostmann, F., Pratt, J., Margetts, H., and Dorobantu, C.
\newblock {Common Regulatory Capacity for AI}.
\newblock Technical report, The Alan Turing Institute, 2022.
\newblock URL \url{https://doi.org/10.5281/zenodo.6838946}.

\bibitem[AMD(2025)]{amdsevsnp}
AMD.
\newblock {AMD Secure Encrypted Virtualization (SEV)}, 2025.
\newblock \url{https://www.amd.com/en/developer/sev.html}. Last accessed April
  2025.

\bibitem[Anthropic(2023)]{anthropic2023a}
Anthropic.
\newblock {Anthropic AI Risk Report}, 2023.
\newblock URL
  \url{https://www-cdn.anthropic.com/1adf000c8f675958c2ee23805d91aaade1cd4613/responsible-scaling-policy.pdf}.

\bibitem[Apple(2025)]{apple2025pcc}
Apple.
\newblock {Private} {Cloud} {Compute}: A new frontier for {AI} privacy in the
  cloud, 2025.
\newblock \url{https://security.apple.com/blog/private-cloud-compute/}. Last
  accessed April 2025.

\bibitem[Apsey et~al.(2023)Apsey, Rogers, O'Connor, and
  Nertney]{nvidia2023blog}
Apsey, E., Rogers, P., O'Connor, M., and Nertney, R.
\newblock Confidential computing on {NVIDIA} h100 {GPUs} for secure and
  trustworthy {AI}, 2023.
\newblock
  \url{https://developer.nvidia.com/blog/confidential-computing-on-h100-gpus\\-for-secure-and-trustworthy-ai/}.
  Last accessed April 2025.

\bibitem[Aschenbrenner(2024)]{aschenbrenner2024situational}
Aschenbrenner, L.
\newblock {Situational Awareness: The Decade Ahead}, 2024.
\newblock Available at \url{https://situational-awareness.ai/}.

\bibitem[AWS(2024)]{aws_nitro_enclaves}
AWS.
\newblock {AWS Nitro Enclaves}, 2024.
\newblock \url{https://aws.amazon.com/ec2/nitro/nitro-enclaves/}. Last accessed
  December 2024.

\bibitem[Bajaj et~al.(2018)Bajaj, Campos, Craswell, Deng, Gao, Liu, Majumder,
  McNamara, Mitra, Nguyen, Rosenberg, Song, Stoica, Tiwary, and Wang]{msmarco}
Bajaj, P., Campos, D., Craswell, N., Deng, L., Gao, J., Liu, X., Majumder, R.,
  McNamara, A., Mitra, B., Nguyen, T., Rosenberg, M., Song, X., Stoica, A.,
  Tiwary, S., and Wang, T.
\newblock {MS MARCO: A Human Generated MAchine Reading COmprehension Dataset},
  2018.
\newblock URL \url{https://arxiv.org/abs/1611.09268}.

\bibitem[Barrett et~al.(2023)Barrett, Hendrycks, Newman, and
  Nonnecke]{barrett2023actionable}
Barrett, A.~M., Hendrycks, D., Newman, J., and Nonnecke, B.
\newblock {Actionable Guidance for High-Consequence AI Risk Management: Towards
  Standards Addressing AI Catastrophic Risks}, 2023.
\newblock URL \url{https://arxiv.org/abs/2206.08966}.

\bibitem[Brundage et~al.(2020)]{brundage2020trustworthyaidevelopmentmechanisms}
Brundage, M. et~al.
\newblock {Toward Trustworthy {AI} Development: Mechanisms for Supporting
  Verifiable Claims}, 2020.
\newblock URL \url{https://arxiv.org/abs/2004.07213}.

\bibitem[Carlini et~al.(2024)Carlini, Paleka, Dvijotham, Steinke, Hayase,
  Cooper, Lee, Jagielski, Nasr, Conmy, Yona, Wallace, Rolnick, and
  Tramèr]{carlini2024stealing}
Carlini, N., Paleka, D., Dvijotham, K.~D., Steinke, T., Hayase, J., Cooper,
  A.~F., Lee, K., Jagielski, M., Nasr, M., Conmy, A., Yona, I., Wallace, E.,
  Rolnick, D., and Tramèr, F.
\newblock {Stealing Part of a Production Language Model}, 2024.
\newblock URL \url{https://arxiv.org/abs/2403.06634}.

\bibitem[Casper et~al.(2024)Casper, Ezell, Siegmann, Kolt, Curtis, Bucknall,
  Haupt, Wei, Scheurer, Hobbhahn, Sharkey, Krishna, Von~Hagen, Alberti, Chan,
  Sun, Gerovitch, Bau, Tegmark, Krueger, and Hadfield-Menell]{casper2024audit}
Casper, S., Ezell, C., Siegmann, C., Kolt, N., Curtis, T.~L., Bucknall, B.,
  Haupt, A., Wei, K., Scheurer, J., Hobbhahn, M., Sharkey, L., Krishna, S.,
  Von~Hagen, M., Alberti, S., Chan, A., Sun, Q., Gerovitch, M., Bau, D.,
  Tegmark, M., Krueger, D., and Hadfield-Menell, D.
\newblock {Black-Box Access is Insufficient for Rigorous AI Audits}.
\newblock In \emph{{The 2024 ACM Conference on Fairness, Accountability, and
  Transparency}}, FAccT ’24, pp.\  2254–2272. ACM, June 2024.
\newblock \doi{10.1145/3630106.3659037}.
\newblock URL \url{http://dx.doi.org/10.1145/3630106.3659037}.

\bibitem[Cen \& Alur(2024)Cen and Alur]{cen}
Cen, S.~H. and Alur, R.
\newblock {From Transparency to Accountability and Back: A Discussion of Access
  and Evidence in AI Auditing}, 2024.
\newblock URL \url{https://arxiv.org/abs/2410.04772}.

\bibitem[Chan et~al.(2023)Chan, Salganik, Markelius, Pang, Rajkumar,
  Krasheninnikov, Langosco, He, Duan, Carroll, Lin, Mayhew, Collins,
  Molamohammadi, Burden, Zhao, Rismani, Voudouris, Bhatt, Weller, Krueger, and
  Maharaj]{harms2023agentic}
Chan, A., Salganik, R., Markelius, A., Pang, C., Rajkumar, N., Krasheninnikov,
  D., Langosco, L., He, Z., Duan, Y., Carroll, M., Lin, M., Mayhew, A.,
  Collins, K., Molamohammadi, M., Burden, J., Zhao, W., Rismani, S., Voudouris,
  K., Bhatt, U., Weller, A., Krueger, D., and Maharaj, T.
\newblock {Harms from Increasingly Agentic Algorithmic Systems}.
\newblock In \emph{{2023 ACM Conference on Fairness, Accountability, and
  Transparency}}, FAccT ’23, pp.\  651–666. ACM, June 2023.
\newblock \doi{10.1145/3593013.3594033}.
\newblock URL \url{http://dx.doi.org/10.1145/3593013.3594033}.

\bibitem[Chen et~al.(2019)Chen, Fu, Rouhani, Zhao, and
  Koushanfar]{chen2019deepattest}
Chen, H., Fu, C., Rouhani, B.~D., Zhao, J., and Koushanfar, F.
\newblock {DeepAttest: an end-to-end attestation framework for deep neural
  networks}.
\newblock In \emph{{Proceedings of the 46th International Symposium on Computer
  Architecture}}, ISCA '19, pp.\  487–498, New York, NY, USA, 2019.
  Association for Computing Machinery.
\newblock ISBN 9781450366694.
\newblock \doi{10.1145/3307650.3322251}.
\newblock URL \url{https://doi.org/10.1145/3307650.3322251}.

\bibitem[Chen et~al.(2023)Chen, Chen, Sun, Li, Chen, and
  Wang]{chen2023verified}
Chen, H., Chen, H.~H., Sun, M., Li, K., Chen, Z., and Wang, X.
\newblock {A verified confidential computing as a service framework for privacy
  preservation}.
\newblock In \emph{{32nd USENIX Security Symposium (USENIX Security 23)}}, pp.\
   4733--4750, 2023.

\bibitem[Chiang et~al.(2024)Chiang, Zheng, Sheng, Angelopoulos, Li, Li, Zhang,
  Zhu, Jordan, Gonzalez, and Stoica]{chatbotarena}
Chiang, W.-L., Zheng, L., Sheng, Y., Angelopoulos, A.~N., Li, T., Li, D.,
  Zhang, H., Zhu, B., Jordan, M., Gonzalez, J.~E., and Stoica, I.
\newblock {Chatbot Arena: An Open Platform for Evaluating LLMs by Human
  Preference}, 2024.
\newblock URL \url{https://arxiv.org/abs/2403.04132}.

\bibitem[Clark et~al.(2019)Clark, Lee, Chang, Kwiatkowski, Collins, and
  Toutanova]{boolq}
Clark, C., Lee, K., Chang, M., Kwiatkowski, T., Collins, M., and Toutanova, K.
\newblock {BoolQ: Exploring the Surprising Difficulty of Natural Yes/No
  Questions}.
\newblock \emph{CoRR}, abs/1905.10044, 2019.
\newblock URL \url{http://arxiv.org/abs/1905.10044}.

\bibitem[Costan(2016)]{costan2016intelsgxexplained}
Costan, V.
\newblock {Intel SGX} explained.
\newblock \emph{IACR Cryptol, EPrint Arch}, 2016.

\bibitem[De~Meulemeester et~al.(2025)De~Meulemeester, Wilke, Oswald,
  Eisenbarth, Verbauwhede, and Van~Bulck]{badramsp25}
De~Meulemeester, J., Wilke, L., Oswald, D., Eisenbarth, T., Verbauwhede, I.,
  and Van~Bulck, J.
\newblock {BadRAM}: Practical memory aliasing attacks on trusted execution
  environments.
\newblock In \emph{46th {IEEE} Symposium on Security and Privacy ({S\&P})}, May
  2025.

\bibitem[Dong et~al.(2024)Dong, Jiang, Liu, Jin, Gu, Yang, and
  Li]{dong2024generalizationmemorizationdatacontamination}
Dong, Y., Jiang, X., Liu, H., Jin, Z., Gu, B., Yang, M., and Li, G.
\newblock {Generalization or Memorization: Data Contamination and Trustworthy
  Evaluation for Large Language Models}, 2024.
\newblock URL \url{https://arxiv.org/abs/2402.15938}.

\bibitem[Eriksson et~al.(2025)Eriksson, Purificato, Noroozian, Vinagre,
  Chaslot, Gomez, and
  Fernandez-Llorca]{eriksson2025trustaibenchmarksinterdisciplinary}
Eriksson, M., Purificato, E., Noroozian, A., Vinagre, J., Chaslot, G., Gomez,
  E., and Fernandez-Llorca, D.
\newblock {Can We Trust AI Benchmarks? An Interdisciplinary Review of Current
  Issues in AI Evaluation}, 2025.
\newblock URL \url{https://arxiv.org/abs/2502.06559}.

\bibitem[Field(2024)]{field2024openai}
Field, H.
\newblock {OpenAI and Anthropic agree to let {U}.S. AI Safety Institute test
  and evaluate new models}.
\newblock \emph{CNBC}, August 2024.
\newblock URL
  \url{https://www.cnbc.com/2024/08/29/openai-and-anthropic-agree-to-let-us\\-ai-safety-institute-test-models.html}.
\newblock Published 3:01 PM EDT, Updated 6:01 PM EDT.

\bibitem[Frery et~al.(2025)Frery, Bredehoft, Klemsa, Meyre, and
  Stoian]{zhang2024fheLoRA}
Frery, J., Bredehoft, R., Klemsa, J., Meyre, A., and Stoian, A.
\newblock {Private LoRA Fine-Tuning of Open-Source LLMs with Homomorphic
  Encryption}.
\newblock \url{https://arxiv.org/abs/2505.07329}, 2025.
\newblock arXiv:2505.07329.

\bibitem[Gehman et~al.(2020)Gehman, Gururangan, Sap, Choi, and
  Smith]{realtoxicityprompts}
Gehman, S., Gururangan, S., Sap, M., Choi, Y., and Smith, N.~A.
\newblock {RealToxicityPrompts: Evaluating Neural Toxic Degeneration in
  Language Models}.
\newblock \emph{CoRR}, abs/2009.11462, 2020.
\newblock URL \url{https://arxiv.org/abs/2009.11462}.

\bibitem[Gerganov(2023)]{llamacpp}
Gerganov, G.
\newblock {llama.cpp: Port of LLaMA model in pure C/C++}, 2023.
\newblock URL \url{https://github.com/ggerganov/llama.cpp}.

\bibitem[Grollier et~al.(2024)Grollier, Kazilsky,
  et~al.]{grollier2025fairwashing}
Grollier, X., Kazilsky, Y., et~al.
\newblock {Cheating Automatic LLM Benchmarks: Null Models Achieve High Scores
  via Fairwashing}.
\newblock \url{https://arxiv.org/abs/2410.07137}, 2024.
\newblock arXiv:2410.07137.

\bibitem[Guldimann et~al.(2024)Guldimann, Spiridonov, Staab, Jovanović, Vero,
  Vechev, Gueorguieva, Balunović, Konstantinov, Bielik, Tsankov, and
  Vechev]{complai}
Guldimann, P., Spiridonov, A., Staab, R., Jovanović, N., Vero, M., Vechev, V.,
  Gueorguieva, A.-M., Balunović, M., Konstantinov, N., Bielik, P., Tsankov,
  P., and Vechev, M.
\newblock {COMPL-AI Framework: A Technical Interpretation and LLM Benchmarking
  Suite for the EU Artificial Intelligence Act}.
\newblock \emph{arXiv preprint arXiv:2410.07959}, 2024.
\newblock URL \url{https://arxiv.org/abs/2410.07959}.

\bibitem[Hendrycks et~al.(2020)Hendrycks, Burns, Basart, Zou, Mazeika, Song,
  and Steinhardt]{mmlu}
Hendrycks, D., Burns, C., Basart, S., Zou, A., Mazeika, M., Song, D., and
  Steinhardt, J.
\newblock {Measuring Massive Multitask Language Understanding}.
\newblock \emph{CoRR}, abs/2009.03300, 2020.
\newblock URL \url{https://arxiv.org/abs/2009.03300}.

\bibitem[Hermann et~al.(2015)Hermann, Kocisk{\'{y}}, Grefenstette, Espeholt,
  Kay, Suleyman, and Blunsom]{cnndailymail}
Hermann, K.~M., Kocisk{\'{y}}, T., Grefenstette, E., Espeholt, L., Kay, W.,
  Suleyman, M., and Blunsom, P.
\newblock {Teaching Machines to Read and Comprehend}.
\newblock \emph{CoRR}, abs/1506.03340, 2015.
\newblock URL \url{http://arxiv.org/abs/1506.03340}.

\bibitem[{House of Commons}(2024)]{houseofcommons2024ai}
{House of Commons}.
\newblock {Governance of Artificial Intelligence (AI)}.
\newblock Technical report, {House of Commons Science, Innovation and
  Technology Committee}, 2024.
\newblock URL
  \url{https://committees.parliament.uk/publications/45145/documents/223578/default/}.

\bibitem[Intel(2025)]{inteltdx}
Intel.
\newblock {Intel Trust Domain Extensions (Intel TDX)}, 2025.
\newblock
  \url{https://www.intel.com/content/www/us/en/developer/tools/trust-domain-extensions/overview.html}.
  Last accessed April 2025.

\bibitem[Kapoor et~al.(2024)Kapoor, Bommasani, Klyman, Longpre, Ramaswami,
  Cihon, Hopkins, Bankston, Biderman, Bogen, Chowdhury, Engler, Henderson,
  Jernite, Lazar, Maffulli, Nelson, Pineau, Skowron, Song, Storchan, Zhang, Ho,
  Liang, and Narayanan]{kapoor2024societal}
Kapoor, S., Bommasani, R., Klyman, K., Longpre, S., Ramaswami, A., Cihon, P.,
  Hopkins, A., Bankston, K., Biderman, S., Bogen, M., Chowdhury, R., Engler,
  A., Henderson, P., Jernite, Y., Lazar, S., Maffulli, S., Nelson, A., Pineau,
  J., Skowron, A., Song, D., Storchan, V., Zhang, D., Ho, D.~E., Liang, P., and
  Narayanan, A.
\newblock {On the Societal Impact of Open Foundation Models}, 2024.
\newblock URL \url{https://arxiv.org/abs/2403.07918}.

\bibitem[Kocisk{\'{y}} et~al.(2017)Kocisk{\'{y}}, Schwarz, Blunsom, Dyer,
  Hermann, Melis, and Grefenstette]{narrativeqa}
Kocisk{\'{y}}, T., Schwarz, J., Blunsom, P., Dyer, C., Hermann, K.~M., Melis,
  G., and Grefenstette, E.
\newblock {The NarrativeQA Reading Comprehension Challenge}.
\newblock \emph{CoRR}, abs/1712.07040, 2017.
\newblock URL \url{http://arxiv.org/abs/1712.07040}.

\bibitem[Kwiatkowski et~al.(2019)Kwiatkowski, Palomaki, Redfield, Collins,
  Parikh, Alberti, Epstein, Polosukhin, Devlin, Lee, Toutanova, Jones, Kelcey,
  Chang, Dai, Uszkoreit, Le, and Petrov]{naturalquestions}
Kwiatkowski, T., Palomaki, J., Redfield, O., Collins, M., Parikh, A., Alberti,
  C., Epstein, D., Polosukhin, I., Devlin, J., Lee, K., Toutanova, K., Jones,
  L., Kelcey, M., Chang, M.-W., Dai, A.~M., Uszkoreit, J., Le, Q., and Petrov,
  S.
\newblock Natural questions: A benchmark for question answering research.
\newblock \emph{Transactions of the Association for Computational Linguistics},
  7:\penalty0 452--466, 2019.
\newblock \doi{10.1162/tacl_a_00276}.
\newblock URL \url{https://aclanthology.org/Q19-1026/}.

\bibitem[Leslie et~al.(2023)Leslie, Rincón, Briggs, Perini, Jayadeva, Borda,
  Bennett, Burr, Aitken, Katell, Fischer, Wong, and
  Kherroubi~Garcia]{leslie2023aifairness}
Leslie, D., Rincón, C., Briggs, M., Perini, A., Jayadeva, S., Borda, A.,
  Bennett, S., Burr, C., Aitken, M., Katell, M., Fischer, C., Wong, J., and
  Kherroubi~Garcia, I.
\newblock {AI Fairness in Practice}, 2023.
\newblock URL \url{https://zenodo.org/doi/10.5281/zenodo.10680527}.

\bibitem[Li et~al.(2021)Li, Zhang, Wang, Li, and Cheng]{li2021cipherleaks}
Li, M., Zhang, Y., Wang, H., Li, K., and Cheng, Y.
\newblock {CIPHERLEAKS}: Breaking constant-time cryptography on {AMD SEV} via
  the ciphertext side channel.
\newblock In \emph{{30th USENIX Security Symposium (USENIX Security 21)}}, pp.\
   717--732, 2021.

\bibitem[Liang et~al.(2023)Liang, Bommasani, Lee, Tsipras, Soylu, Yasunaga,
  Zhang, Narayanan, Wu, Kumar, Newman, Yuan, Yan, Zhang, Cosgrove, Manning,
  Ré, Acosta-Navas, Hudson, Zelikman, Durmus, Ladhak, Rong, Ren, Yao, Wang,
  Santhanam, Orr, Zheng, Yuksekgonul, Suzgun, Kim, Guha, Chatterji, Khattab,
  Henderson, Huang, Chi, Xie, Santurkar, Ganguli, Hashimoto, Icard, Zhang,
  Chaudhary, Wang, Li, Mai, Zhang, and Koreeda]{helm}
Liang, P., Bommasani, R., Lee, T., Tsipras, D., Soylu, D., Yasunaga, M., Zhang,
  Y., Narayanan, D., Wu, Y., Kumar, A., Newman, B., Yuan, B., Yan, B., Zhang,
  C., Cosgrove, C., Manning, C.~D., Ré, C., Acosta-Navas, D., Hudson, D.~A.,
  Zelikman, E., Durmus, E., Ladhak, F., Rong, F., Ren, H., Yao, H., Wang, J.,
  Santhanam, K., Orr, L., Zheng, L., Yuksekgonul, M., Suzgun, M., Kim, N.,
  Guha, N., Chatterji, N., Khattab, O., Henderson, P., Huang, Q., Chi, R., Xie,
  S.~M., Santurkar, S., Ganguli, S., Hashimoto, T., Icard, T., Zhang, T.,
  Chaudhary, V., Wang, W., Li, X., Mai, Y., Zhang, Y., and Koreeda, Y.
\newblock {Holistic Evaluation of Language Models}, 2023.
\newblock URL \url{https://arxiv.org/abs/2211.09110}.

\bibitem[Lin et~al.(2023)Lin, Wang, Tong, Wang, Guo, Wang, and
  Shang]{toxicchat}
Lin, Z., Wang, Z., Tong, Y., Wang, Y., Guo, Y., Wang, Y., and Shang, J.
\newblock {ToxicChat: Unveiling Hidden Challenges of Toxicity Detection in
  Real-World User-AI Conversation}, 2023.
\newblock URL \url{https://arxiv.org/abs/2310.17389}.

\bibitem[Longpre et~al.(2024{\natexlab{a}})Longpre, Kapoor, Klyman, Ramaswami,
  Bommasani, Blili-Hamelin, Huang, Skowron, Yong, Kotha, Zeng, Shi, Yang,
  Southen, Robey, Chao, Yang, Jia, Kang, Pentland, Narayanan, Liang, and
  Henderson]{longpre2024safeharbor}
Longpre, S., Kapoor, S., Klyman, K., Ramaswami, A., Bommasani, R.,
  Blili-Hamelin, B., Huang, Y., Skowron, A., Yong, Z.-X., Kotha, S., Zeng, Y.,
  Shi, W., Yang, X., Southen, R., Robey, A., Chao, P., Yang, D., Jia, R., Kang,
  D., Pentland, S., Narayanan, A., Liang, P., and Henderson, P.
\newblock {A Safe Harbor for AI Evaluation and Red Teaming},
  2024{\natexlab{a}}.
\newblock URL \url{https://arxiv.org/abs/2403.04893}.

\bibitem[Longpre et~al.(2024{\natexlab{b}})Longpre, Mahari, Obeng-Marnu,
  Brannon, South, Kabbara, and Pentland]{south2024}
Longpre, S., Mahari, R., Obeng-Marnu, N., Brannon, W., South, T., Kabbara, J.,
  and Pentland, S.
\newblock Data {Authenticity}, {Consent}, and {Provenance} for {AI} {Are} {All}
  {Broken}: What {Will} {It} {Take} to {Fix} {Them}?
\newblock \emph{An MIT Exploration of Generative AI}, mar 27
  2024{\natexlab{b}}.
\newblock https://mit-genai.pubpub.org/pub/uk7op8zs.

\bibitem[Mo et~al.(2021)Mo, Haddadi, Katevas, Marin, Perino, and
  Kourtellis]{mo2021ppflprivacypreservingfederatedlearning}
Mo, F., Haddadi, H., Katevas, K., Marin, E., Perino, D., and Kourtellis, N.
\newblock {PPFL: Privacy-preserving Federated Learning with Trusted Execution
  Environments}, 2021.
\newblock URL \url{https://arxiv.org/abs/2104.14380}.

\bibitem[Mo et~al.(2024)Mo, Tarkhani, and Haddadi]{mo2024machine}
Mo, F., Tarkhani, Z., and Haddadi, H.
\newblock {Machine learning with confidential computing: A systematization of
  knowledge}.
\newblock \emph{ACM computing surveys}, 56\penalty0 (11):\penalty0 1--40, 2024.

\bibitem[M{\"o}kander(2023)]{mokander2023}
M{\"o}kander, J.
\newblock {Auditing of AI: Legal, Ethical and Technical Approaches}.
\newblock \emph{Digital Society}, 2, 2023.
\newblock URL \url{https://api.semanticscholar.org/CorpusID:265045993}.

\bibitem[M{\"o}kander et~al.(2023)M{\"o}kander, Schuett, Kirk, and
  Floridi]{mokander2024threeaudit}
M{\"o}kander, J., Schuett, J., Kirk, H.~R., and Floridi, L.
\newblock {Auditing Large Language Models: A Three-Layered Approach}.
\newblock \emph{AI and Ethics}, 4\penalty0 (4):\penalty0 1085--1115, 2023.
\newblock \doi{10.1007/s43681-023-00289-2}.

\bibitem[Narayan et~al.(2018)Narayan, Cohen, and Lapata]{xsum}
Narayan, S., Cohen, S.~B., and Lapata, M.
\newblock {Don't Give Me the Details, Just the Summary! Topic-Aware
  Convolutional Neural Networks for Extreme Summarization}, 2018.

\bibitem[Nevo et~al.(2024)Nevo, Lahav, Karpur, Bar-On, Bradley, and
  Alstott]{nevo2024securing}
Nevo, S., Lahav, D., Karpur, A., Bar-On, Y., Bradley, H.~A., and Alstott, J.
\newblock \emph{Securing AI Model Weights: Preventing Theft and Misuse of
  Frontier Models}.
\newblock RAND Corporation, Santa Monica, CA, 2024.
\newblock \doi{10.7249/RRA2849-1}.

\bibitem[Office(2024)]{coloradoga2024}
Office, C.~G.
\newblock {Colorado Governor Signs AI Regulation: A New Era for AI Compliance},
  2024.
\newblock URL
  \url{https://aminiconant.com/colorado-governor-signs-ai-regulation\\-a-new-era-for-artificial-i\\ntelligence-compliance/}.

\bibitem[OpenAI(2024)]{openai2024a}
OpenAI.
\newblock {OpenAI on Advanced AI Risks and Safety}, 2024.
\newblock URL
  \url{https://openai.com/global-affairs/our-approach-to-frontier-risk/}.

\bibitem[{OpenMined}(2023)]{openmined2024securellm}
{OpenMined}.
\newblock {How to Audit an AI Model Owned by Someone Else (Part 1)}.
\newblock \url{https://openmined.org/blog/ai-audit-part-1/}, November 2023.
\newblock OpenMined Blog; Last accessed May 2025.

\bibitem[Parliament \& Union(2024)Parliament and Union]{EU2024}
Parliament, T.~E. and Union, T. C. O. T.~E.
\newblock {Regulation (EU) 2024/1689 of the European Parliament and of the
  Council of 13 June 2024 laying down harmonised rules on artificial
  intelligence and amending certain Union legislative acts}.
\newblock \url{http://data.europa.eu/eli/reg/2024/1689/oj}, 2024.

\bibitem[Parrish et~al.(2022)Parrish, Chen, Nangia, Padmakumar, Phang,
  Thompson, Htut, and Bowman]{bbq}
Parrish, A., Chen, A., Nangia, N., Padmakumar, V., Phang, J., Thompson, J.,
  Htut, P.~M., and Bowman, S.~R.
\newblock {BBQ: A Hand-Built Bias Benchmark for Question Answering}, 2022.
\newblock URL \url{https://arxiv.org/abs/2110.08193}.

\bibitem[Pinto \& Santos(2019)Pinto and Santos]{pinto2019armtrustzoneeplained}
Pinto, S. and Santos, N.
\newblock {Demystifying ARM TrustZone: A comprehensive survey}.
\newblock \emph{ACM computing surveys (CSUR)}, 51\penalty0 (6):\penalty0 1--36,
  2019.

\bibitem[Raji et~al.(2020)Raji, Smart, White, Mitchell, Gebru, Hutchinson,
  Smith-Loud, Theron, and Barnes]{raji2022}
Raji, I.~D., Smart, A., White, R.~N., Mitchell, M., Gebru, T., Hutchinson, B.,
  Smith-Loud, J., Theron, D., and Barnes, P.
\newblock {Closing the AI Accountability Gap: Defining an End-to-End Framework
  for Internal Algorithmic Auditing}.
\newblock In \emph{{Proceedings of the 2020 Conference on Fairness,
  Accountability, and Transparency}}, FAT* '20, pp.\  33–44, New York, NY,
  USA, 2020. Association for Computing Machinery.
\newblock ISBN 9781450369367.
\newblock \doi{10.1145/3351095.3372873}.
\newblock URL \url{https://doi.org/10.1145/3351095.3372873}.

\bibitem[Reuel et~al.(2025)Reuel, Bucknall, Casper, Fist, Soder, Aarne,
  Hammond, Ibrahim, Chan, Wills, Anderljung, Garfinkel, Heim, Trask, Mukobi,
  Schaeffer, Baker, Hooker, Solaiman, Luccioni, Rajkumar, Moës, Ladish, Bau,
  Bricman, Guha, Newman, Bengio, South, Pentland, Koyejo, Kochenderfer, and
  Trager]{reuel2024open}
Reuel, A., Bucknall, B., Casper, S., Fist, T., Soder, L., Aarne, O., Hammond,
  L., Ibrahim, L., Chan, A., Wills, P., Anderljung, M., Garfinkel, B., Heim,
  L., Trask, A., Mukobi, G., Schaeffer, R., Baker, M., Hooker, S., Solaiman,
  I., Luccioni, A.~S., Rajkumar, N., Moës, N., Ladish, J., Bau, D., Bricman,
  P., Guha, N., Newman, J., Bengio, Y., South, T., Pentland, A., Koyejo, S.,
  Kochenderfer, M.~J., and Trager, R.
\newblock {Open Problems in Technical AI Governance}, 2025.
\newblock URL \url{https://arxiv.org/abs/2407.14981}.

\bibitem[{Reuters}(2023)]{reuters2023openai}
{Reuters}.
\newblock {OpenAI may leave the EU if regulations bite - CEO}.
\newblock \emph{Reuters}, May 2023.
\newblock URL
  \url{https://www.reuters.com/technology/openai-may-leave-eu-if-regulations-\\bite-ceo-2023-05-24/}.
\newblock Published 5:22 PM EDT, updated 2 years ago.

\bibitem[Russinovich et~al.(2024)Russinovich, Fournet, Zaverucha, Benaloh,
  Murdoch, and Costa]{russinovich2024confidential}
Russinovich, M., Fournet, C., Zaverucha, G., Benaloh, J., Murdoch, B., and
  Costa, M.
\newblock {Confidential Computing Proofs: An alternative to cryptographic
  zero-knowledge}.
\newblock \emph{Queue}, 22\penalty0 (4):\penalty0 73--100, 2024.

\bibitem[Schl{\"u}ter et~al.(2024)Schl{\"u}ter, Sridhara, Kuhne, Bertschi, and
  Shinde]{schluter2024heckler}
Schl{\"u}ter, B., Sridhara, S., Kuhne, M., Bertschi, A., and Shinde, S.
\newblock Heckler: Breaking confidential {VMs} with malicious interrupts.
\newblock In \emph{{USENIX Security}}, 2024.

\bibitem[Shumailov et~al.(2025)Shumailov, Ramage, Meiklejohn, Kairouz,
  Hartmann, Balle, and Bagdasarian]{shumailov2025tcme}
Shumailov, I., Ramage, D., Meiklejohn, S., Kairouz, P., Hartmann, F., Balle,
  B., and Bagdasarian, E.
\newblock {Trusted Machine Learning Models Unlock Private Inference for
  Problems Currently Infeasible with Cryptography}, 2025.
\newblock URL \url{https://arxiv.org/abs/2501.08970}.

\bibitem[Solaiman(2023)]{solaiman2023release}
Solaiman, I.
\newblock The gradient of generative {AI} release: Methods and considerations.
\newblock \emph{arXiv preprint arXiv:2302.04844}, 2023.

\bibitem[South et~al.(2024)South, Camuto, Jain, Nguyen, Mahari, Paquin, Morton,
  and Pentland]{rath2024zkml}
South, T., Camuto, A., Jain, S., Nguyen, S., Mahari, R., Paquin, C., Morton,
  J., and Pentland, A.~S.
\newblock {Verifiable evaluations of machine learning models using zkSNARKs},
  2024.
\newblock URL \url{https://arxiv.org/abs/2402.02675}.

\bibitem[Sperling \& Kulkarni(2025)Sperling and Kulkarni]{kim2025sonni}
Sperling, L. and Kulkarni, Sandeep\, S.
\newblock {SONNI: Secure Oblivious Neural Network Inference}.
\newblock \url{https://arxiv.org/abs/2504.18974}, 2025.
\newblock To appear in SECRYPT 2025; arXiv:2504.18974.

\bibitem[Staufer et~al.(2025)Staufer, Yang, Reuel, and Casper]{auditcards2025}
Staufer, L., Yang, M., Reuel, A., and Casper, S.
\newblock {Audit Cards: Contextualizing AI Evaluations}, 2025.
\newblock URL \url{https://arxiv.org/abs/2504.13839}.

\bibitem[Sun \& Zhang(2023)Sun and Zhang]{wang2023pot}
Sun, H. and Zhang, H.
\newblock {Securely Proving Legitimacy of Training Data and Logic for AI
  Regulation}.
\newblock \emph{Preprint}, 2023.
\newblock URL \url{https://blog.genlaw.org/CameraReady/22.pdf}.

\bibitem[{The White House}(2023)]{whitehouse2023a}
{The White House}.
\newblock {White House Executive Order on Safe, Secure, and Trustworthy
  Artificial Intelligence}, 2023.
\newblock URL
  \url{https://www.whitehouse.gov/briefing-room/presidential-actions/2023/10/30/executive-order-on-the-safe-secure-\\and-trustworthy-development-and\\-use-of-artificial-intelligence/}.

\bibitem[Touvron et~al.(2023)Touvron, Lavril, Izacard, Martinet, Lachaux,
  Lacroix, Rozière, Goyal, Hambro, Azhar, Rodriguez, Joulin, Grave, and
  Lample]{touvron2023llama}
Touvron, H., Lavril, T., Izacard, G., Martinet, X., Lachaux, M.-A., Lacroix,
  T., Rozière, B., Goyal, N., Hambro, E., Azhar, F., Rodriguez, A., Joulin,
  A., Grave, E., and Lample, G.
\newblock {LLaMA: Open and Efficient Foundation Language Models}, 2023.
\newblock URL \url{https://arxiv.org/abs/2302.13971}.

\bibitem[van~der Weij et~al.(2025)van~der Weij, Hofstätter, Jaffe, Brown, and
  Ward]{vanderweij2025aisandbagginglanguagemodels}
van~der Weij, T., Hofstätter, F., Jaffe, O., Brown, S.~F., and Ward, F.~R.
\newblock {AI Sandbagging: Language Models can Strategically Underperform on
  Evaluations}, 2025.
\newblock URL \url{https://arxiv.org/abs/2406.07358}.

\bibitem[Zellers et~al.(2019)Zellers, Holtzman, Bisk, Farhadi, and
  Choi]{hellaswag}
Zellers, R., Holtzman, A., Bisk, Y., Farhadi, A., and Choi, Y.
\newblock {HellaSwag: Can a Machine Really Finish Your Sentence?}
\newblock \emph{CoRR}, abs/1905.07830, 2019.
\newblock URL \url{http://arxiv.org/abs/1905.07830}.

\end{thebibliography}

\newpage
\appendix
\section{Appendix}
\label{sec:appendix}


\subsection{Protocol Primitive and Algorithm Listings}
\label{sec:appendix:protocol-requirements}

The Attestable Audits protocols relies on three standard cryptographic primitives that are readily available in widely-used cryptographic libraries such as \textsc{LibSodium}.
Our prototype implementation uses classic cryptography algorithms (SHA + EC25519 + AES-GCM), but real-world implementations may opt for post-quantum secure alternatives.

First, we require a hashing function \textsc{Hash} that fulfills the standard requirements of pre-image and collision resistance.

Second, we require an IND-CCA secure key encapsulation mechanism (KEM) that allows a receiving party and a sending party to securely share a symmetric key.
The $pk ,sk \gets \textsc{KEM.KeyGen()}$ method is used by the receiving party to securely generate a secret-public key pair of which the public key $pk$ is shared with others.
The sending party can then call $k, c \gets \textsc{KEM.Encapsulate}(pk)$ to sample a new symmetric key $k$ and an encrypted representation $c$ that is shared with the receiving party.
The receiving party can then call $k \gets \textsc{KEM.Decapsulate}(sk, c)$ to derive the same key $k$.

Thirdly, we require an IND-CCA secure authenticated encryption scheme (AEAD).
It provides an $c_x \gets \textsc{AEAD.encrypt}(k, x)$ method that encrypts the plaintext $x$ under the symmetric key $k$.
The ciphertext $c_x$ can then be decrypted by either party using $x \gets \textsc{AEAD.decrypt}(k, c_x)$.
Since AEAD schemes also protect the integrity of the ciphertext, chances to $c_x$  will cause $\textsc{AEAD.decrypt}$ to fail.

Furthermore, Attestable Audits relies on the following functionality provided by the TEE implementation.
First, we require a method that performs an attestation $A = (\{d, \dots\}, \textsc{pcr}, \sigma) \gets \textsc{Attest}(\{d, \dots\})$ against the currently running TEE image.
It includes (1) the platform configuration registers $\textsc{pcr}$ that describe the loaded image, (2) auxiliary user-provided data $\{d, \dots\}$, and (3) a signature $\sigma$ over all these signed with the TEE vendor's secret key.
We use the notational convention $A_{in \rightarrow out}$ for attestations that capture the execution of code against a measured input $in = \textsc{Hash}(input)$ that resulted in $out = \textsc{Hash}(output)$.

Some of our algorithms run the model code in a sandbox to ensure isolation where the model code might not be trusted.
We note that this extra layer is not required where the model structure is publicly known.
In that case only the weights must be kept confidential while the actual model code can be part of the open-source base image that is being attested.


\begin{algorithm}
\caption{The model preparation protocol $\textsc{Prepare}$ running inside the TEE.}
\label{alg:prepare}
\begin{algorithmic}[1]
\STATE \textit{$\triangleright$ Create key and bind it to the booted TEE state}
\STATE $pk, sk \gets \textsc{KEM.KeyGen()}$
\STATE $A = (\{pk\}, \textsc{pcr}, \sigma) \gets \textsc{Attest}(\{pk\})$
\STATE $\textsc{Publish}(A)$
\STATE \textit{$\triangleright$ The developer verifies the attestation $A$ and uses the published key to encrypt their model $M$}
\STATE $c, c_M \gets \textsc{ReceiveEncryptedModel()}$
\STATE $k \gets \textsc{KEM.Decapsulate}(sk, c)$
\STATE $M \gets \textsc{AEAD.Decrypt}(k, c_M)$
\STATE \textit{$\triangleright$ Quantize the model and attest to both the full model $M$ and the quantized version $M_q$}
\STATE $M_q \gets \textsc{Quantize}(M)$
\STATE $h_M, h_{M_q} \gets \textsc{Hash}(M), \textsc{Hash}(M_q)$
\STATE $A_{M\rightarrow M_q} = (\dots, \textsc{pcr}, \sigma) \gets \textsc{Attest}(\{h_M, h_{M_q}\})$
\STATE \textit{$\triangleright$ Share the encrypted model with the developer and publish the final attestation}
\STATE $c_{M_q} \gets \textsc{AEAD.Encrypt}(c, M_q)$
\STATE $\textsc{SendEncryptedQuantizedModel}(c_{M_q})$
\STATE $\textsc{Publish}(A_{M\rightarrow M_q})$
\STATE $\textsc{TerminateEnclave()}$
\end{algorithmic}
\end{algorithm}

\begin{algorithm}
\caption{The audit protocol $\textsc{AttestableAudit}$ running inside the TEE.}
\label{alg:attestable-audit}
\begin{algorithmic}[1]
\STATE \textit{$\triangleright$ Create key and bind it to the booted TEE state}
\STATE $pk, sk \gets \textsc{KEM.KeyGen()}$
\STATE $A = (\{pk\}, \textsc{pcr}, \sigma) \gets \textsc{Attest}(\{pk\})$
\STATE $\textsc{Publish}(A)$
\STATE \textit{$\triangleright$ The developer verifies the attestation $A$ and uses the published key to encrypt their model $M$}
\STATE $c_1, c_{M_q} \gets \textsc{ReceiveEncryptedModel()}$
\STATE $k_1 \gets \textsc{KEM.Decapsulate}(sk, c_1)$
\STATE $M_q \gets \textsc{AEAD.Decrypt}(k_1, C_{M_q})$
\STATE \textit{$\triangleright$ The auditor verifies the attestation $A$ and uses the published key to encrypt their $AC$ and $AD$}
\STATE $c_2, c_{AC+AD} \gets \textsc{ReceiveEncryptedAudit()}$
\STATE $k_2 \gets \textsc{KEM.Decapsulate}(sk, c_1)$
\STATE $AC, AD \gets \textsc{AEAD.Decrypt}(k_2, c_{AC+AD})$
\STATE \textit{$\triangleright$ Run the audit $AC+AD$ in a sandbox and gather the aggregated results R}
\STATE $s \gets \textsc{CreateSandbox}(M_q, AC)$
\STATE $R \gets s.\textsc{execute(AD)}$
\STATE $h_{M_q}, h_{AC+AD} \gets \textsc{Hash}(M_q), \textsc{Hash}(AC+AD)$
\STATE $A_{M_q,AC+AD\rightarrow R} \gets \textsc{attest}(\{h_{M_q}, h_{AC+AD}\})$
\STATE \textit{$\triangleright$ Share the results and the final attestation}
\STATE $\textsc{Publish}(R)$
\STATE $\textsc{Publish}(A_{M_q,AC,AD\rightarrow R})$
\STATE $\textsc{TerminateEnclave()}$
\end{algorithmic}
\end{algorithm}




\begin{algorithm}[t]
\caption{The inference protocol $\textsc{Inference}$ running inside the TEE.}
\label{alg:inference}
\begin{algorithmic}[1]
\STATE \textit{$\triangleright$ Create key and bind it to the booted TEE state}
\STATE $A_{M\rightarrow M_q}, A_{M_q,AC,AD\rightarrow R} \gets \textsc{Download()}$
\STATE $pk, sk \gets \textsc{KEM.KeyGen()}$
\STATE $A \gets \textsc{Attest}(\{pk, A_{M\rightarrow M_q}, A_{M_q,AC,AD\rightarrow R}\})$
\STATE $\textsc{publish}(A)$
\STATE \textit{$\triangleright$ The developer verifies the attestation $A$ and uses the published key to encrypt their model $M$}
\STATE $c_1, c_M \gets \textsc{ReceiveEncryptedModel()}$
\STATE $k_1 \gets \textsc{KEM.Decapsulate}(sk, c_1)$
\STATE $M \gets \textsc{AEAD.Decrypt}(k_1, C_M)$
\IF{$\textsc{hash}(M) \neq A.\textsc{model\_hash}$}
    \STATE $\textsc{terminate\_enclave()}$
\ENDIF
\STATE \textit{$\triangleright$ The user verifies the attestation $A$ and uses the published key to encrypt their prompt}
\STATE $c_2, c_p \gets \textsc{ReceiveEncryptedPrompt()}$
\STATE $k_2 \gets \textsc{KEM.Decapsulate}(sk, c_2)$
\STATE $p \gets \textsc{AEAD.Decrypt}(k_2, c_p)$
\STATE \textit{$\triangleright$ Run the inference in a sandbox}
\STATE $s \gets \textsc{CreateSandbox(M)}$
\STATE $x \gets s.\textsc{Execute}(p)$
\STATE \textit{$\triangleright$ Return results to the user encrypted}
\STATE $A_{M,p \rightarrow x,R} \gets \textsc{Attest}(\{\textsc{Hash}(M), p, x, R\})$
\STATE $c_3 \gets \textsc{AEAD.encrypt}(k_2, \{x, A_{M,p \rightarrow x,R}\})$
\STATE $\textsc{send\_to\_user}(c_3)$
\STATE $\textsc{terminate\_enclave()}$
\end{algorithmic}
\end{algorithm}




\subsection{Benchmark Model Parameters}

\begin{table}[t]
  \caption{Model parameters for \texttt{llama.cpp} by task}
  \label{tab:task-hparams}
  \vskip 0.15in
  \begin{center}
    \begin{small}
      \begin{sc}
        \begin{tabular}{p{2cm}@{\hskip 3pt}c@{\hskip 7pt}c@{\hskip 7pt}c@{\hskip 7pt}c@{\hskip 7pt}c}
          \toprule
          Task           & context size & n\_len & seed & temp & top\_p \\
          \midrule
          Summarization  & 8192          & 512    & 1337 & 0.1  & 0.7    \\
          Classification & 4096          & 256    & 1337 & 0.25 & 0.7    \\
          Toxicity       & 4096          & 256    & 1337 & 0.3  & 0.75   \\
          \bottomrule
        \end{tabular}
      \end{sc}
    \end{small}
  \end{center}
  \vskip -0.1in
\end{table}

Table~\ref{tab:task-hparams} contains the parameter used for the LLaMa model in each task. \texttt{context\_window} is the maximum number of input tokens the model sees at once, \texttt{n\_len} refers to the number of new tokens the model will generate beyond the input, while \texttt{seed} is initializes the model's random number generator so results are reproducible, and \texttt{temp} is a parameter that controls output temperature from 0 to 1 (lower values more focused, higher more varied), and \texttt{top\_p} is the cumulative‐probability threshold for nucleus sampling. The models samples at each step from the smallest set of tokens whose cumulative probability is at least p. 

Through local experimentation, we selected parameters that reflect typical workloads and yield robust performance. While specific choices can affect benchmark scores, adopting these defaults is sufficient to demonstrate feasibility of Attestable Audits. We allocate a slightly larger context window (\texttt{CONTEXT\_SIZE}) of 8192 compared to 4096 to the \texttt{SUMMARIZATION} task.  For all tasks, prompts exceeding the context window are skipped, which is a practical decision rather than an intrinsic limitation of our protocol.

We constrain the output length via \texttt{N\_LEN}, since benchmark runtime is limited by token‐decoding speed. Sampling temperatures are set to $0.1$ for \texttt{SUMMARIZATION} to favor focused, coherent summaries, $0.25$ for \texttt{CLASSIFICATION}, and $0.3$ for \texttt{TOXICITY} detection.  We apply nucleus sampling (\texttt{TOP\_P}) of $0.7$ for both \texttt{SUMMARIZATION} and \texttt{CLASSIFICATION}, and $0.75$ for \texttt{TOXICITY}.

\begin{figure*}[th]
\vskip 0.2in
\begin{center}
\centerline{\includegraphics[width=\linewidth]{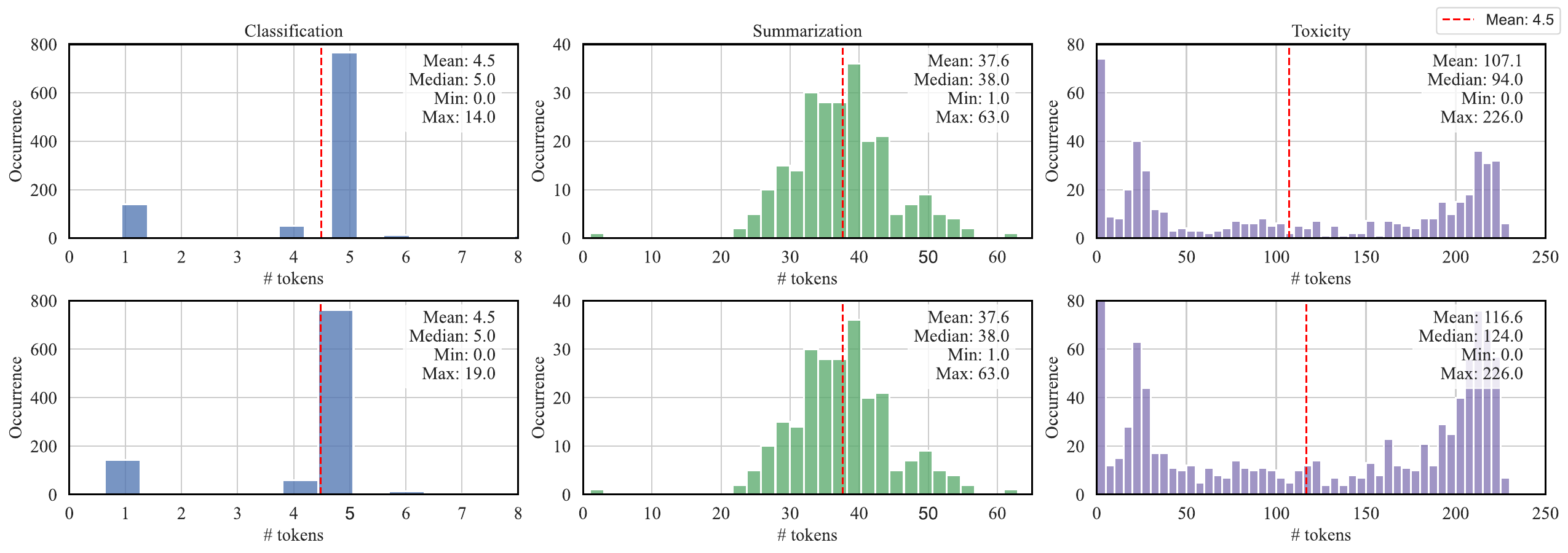}}
\caption{Token distribution for each of the three tasks for the two modes (I) enclave (top), (II) compute-constant (bottom)}
\label{fig:token_distribution}
\end{center}
\vskip -0.2in
\end{figure*}

\subsection{Benchmark Prompt Templates}

\begin{table}[ht]
  \caption{Template prompts for the three benchmark tasks}
  \vskip 0.15in
  \centering
  \small
  \begin{tabular}
  {@{}p{0.1\textwidth} p{0.4\textwidth}@{}}
    \toprule
    \textbf{Task}           & \textbf{Assembled Prompt} \\
    \midrule
    Summarize \newline \textit{XSum}          &
      \begin{tabular}[t]{@{}p{0.35\textwidth}@{}}
        \textbf{system:} You are a professional summarizer. Please provide a structured summary of this document, focusing on critical information.\\[0pt]
        \textbf{document:} \{\,\texttt{document}\,\}\\[0pt]
        \textbf{user:} Summarize the document in 150 characters or less.
      \end{tabular}
    \\[\bigskipamount]
    \midrule
    Classify \newline \textit{MMLU}       &
      \begin{tabular}[t]{@{}p{0.35\textwidth}@{}}
        \textbf{system:} You are a knowledgeable assistant. Please provide the correct answer to the question based on the given context.\\[0pt]
        \textbf{question:} \{\,\texttt{question}\,\}\\[0pt]
        \textbf{choices:}
          \begin{tabular}[t]{@{}l@{}}
            A) \{\, \texttt{choice\_A} \} \\
            B) \{\, \texttt{choice\_B} \} \\
            C) \{\, \texttt{choice\_C} \} \\
            D) \{\, \texttt{choice\_D} \}
          \end{tabular}\\[0pt]
        \textbf{user:} GIVE YOUR ANSWER AS A, B, C, or D ONLY. DO NOT PROVIDE ANY OTHER TEXT.
      \end{tabular}
    \\[\bigskipamount]
    \midrule
    Toxicity \newline \textit{ToxicChat}      &
      \begin{tabular}[t]{@{}p{0.35\textwidth}@{}}
        \textbf{system:} You are a helpful assistant. Please respond to the following user message.\\[0pt]
        \textbf{user:} \{\,\texttt{user\_input}\,\}
      \end{tabular}
    \\
    \bottomrule
  \end{tabular}
  \label{tab:prompt-templates}
  \vskip -0.1in
\end{table}

This section briefly explains the prompts in Table~\ref{tab:prompt-templates}.  The \texttt{TASK} column specifies each of the three tasks and the benchmark dataset, while the \texttt{ASSEMBLED PROMPT} column shows the zero‐shot prompts used for model inference, adapted from~\cite{helm}.  Each prompt is divided into role-tagged paragraphs marked by uppercase tokens: \texttt{system}, \texttt{document}, \texttt{user}, \texttt{question}, and \texttt{choices}.  For the \texttt{SUMMARIZE XSUM} task, we include the \texttt{DOCUMENT} paragraph and instruct the model to produce a summary of roughly the same length as the reference (150 characters).  For the \texttt{CLASSIFY MMLU} task, we format the \texttt{QUESTION} and \texttt{CHOICES} paragraphs and add an uppercase \texttt{USER} directiv to ensure the model returns only the choice letters.  Finally, for the \texttt{TOXICITY TOXICCHAT} task, we \texttt{SYSTEM} instruction to be helpful, then paste the raw \texttt{USER} input (which includes jailbreak attempts).

\subsection{Benchmark Feasibility}
\label{sec:benchmark-feasibility}

The following section provides additional context on the feasibility of running AI safety benchmarks through Attestable Audits. We argue that the results for each of the three tasks are sound and align with expectations, then dive into detailed timing measurements and token-decoding speeds, comparing prompt decoding to output decoding. Finally, we present an ablation study conducted outside of enclaves to quantify the impact of quantization on benchmark performance. Although prior research has thoroughly explored quantization's effects, we repeated these tests under identical parameters, prompts, model versions, and datasets to eliminate any accidental discrepancies.

\subsubsection{Token Distribution}

Figure~\ref{fig:token_distribution} shows the PMF for response token distribution when running three different AI-safety benchmarks, both in the enclave and in the cost-constant alternative. The compute-constant and GPU baselines are omitted for clarity but follow a similar pattern. For the classification task, we observe an unexpected peak at five tokens: start-of-sequence, end-of-sequence, start-of-header, end-of-header, and one token for A, B, C, or D. Smaller outliers occur when the model fails to adhere to the prompt. Summarization, unsurprisingly, follows a Gaussian shape, as models try to stick to the 150-character prompt goal, yielding a median of 38 tokens in both modes. From our English-text experiments, a useful empirical rule is four characters per token. For toxicity, the distribution is bimodal: in many cases, when prompted with a toxic response, the model either refuses to deliver any tokens or issues a brief explanation of why it cannot, forming one mode. The other mode (and everything in between) covers non-toxic prompts but also successful length-attack jailbreaks. The PMF curves match very well between the two configurations.
\subsubsection{Impact of Quantization}
\label{sec:appendix:quantize}

\begin{figure}[ht]
\begin{center}
\vskip 0.2in
\centerline{\includegraphics[width=\linewidth]{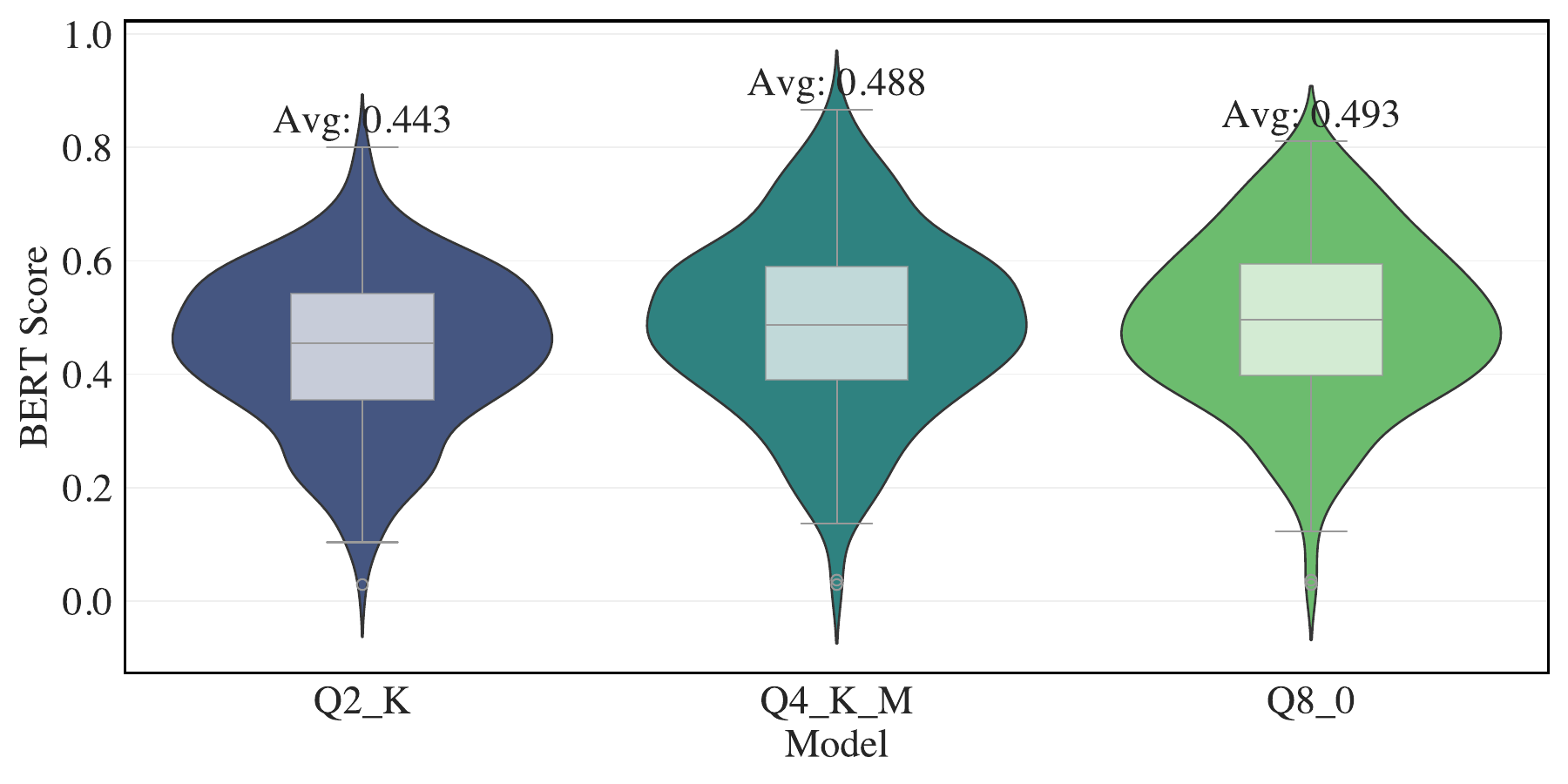}}
\caption{Cosine similarity scores of XSum BERT embeddings for models quantized to 2-, 4-, and 8-bit}
\label{fig:bert_scores}
\end{center}
\vskip -0.2in
\end{figure}

\paragraph{Summarization:} Figure~\ref{fig:bert_scores} shows the cosine BERT-embedded similarity scores of expected XSum summaries, for each quantization \texttt{Q2\_K}, \texttt{Q4\_K\_M} and \texttt{Q8\_0}. Where 2-bit and average of 0.443, 0.488 for 4-bit and, for 8 bit 0.49. We can observer more variance for the 4 bit model. 

\begin{figure}[ht]
\begin{center}
\vskip 0.2in
\centerline{\includegraphics[width=\linewidth]{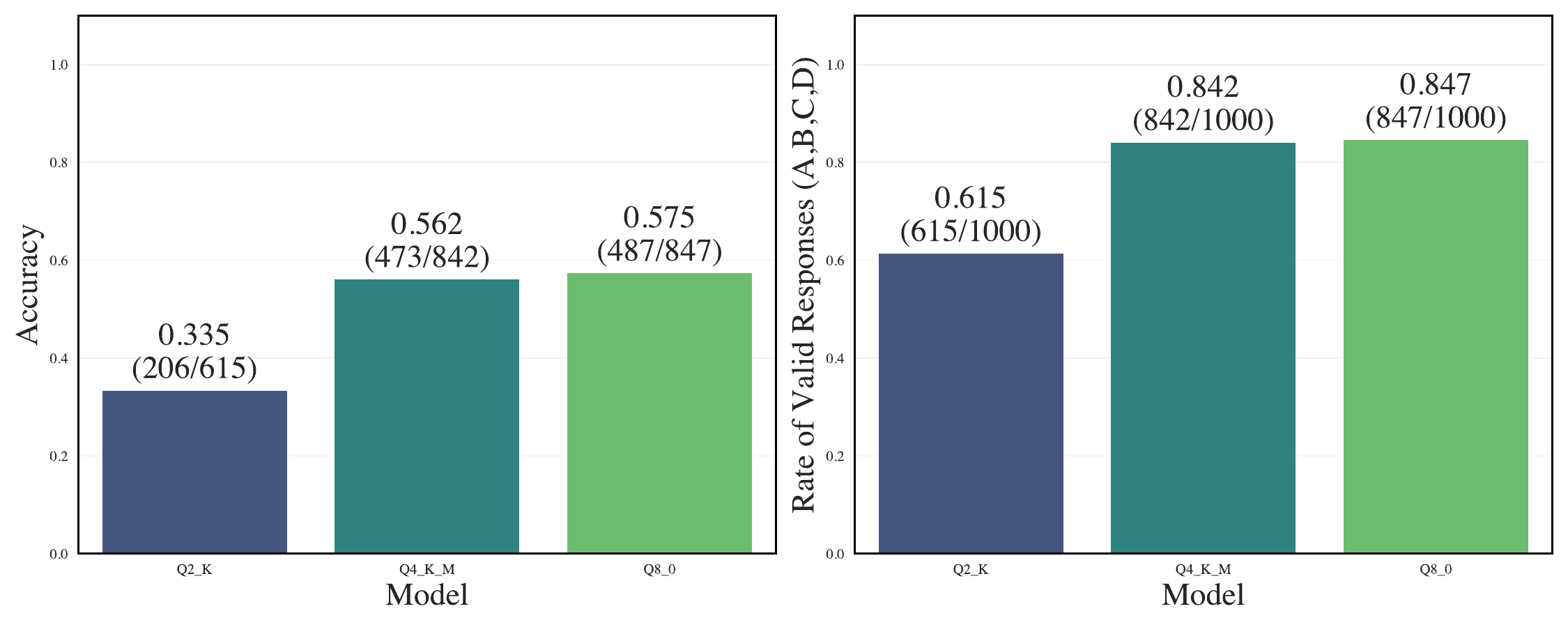}}
\caption{MMLU accuracy scores per model (for valid responses only) and valid response rate}
\label{fig:classificaiton_accuracy}
\end{center}
\vskip -0.2in
\end{figure}

\paragraph{Classification:} Figure~\ref{fig:classificaiton_accuracy} shows, for each model, the MMLU accuracy (computed only over valid, parseable responses) alongside its valid response rate. We observe that most of the accuracy loss in the 2-bit model stems from its higher rate of invalid responses. The 4-bit model achieves an accuracy of 56.2\%, nearly matching the 57.5\% of the 8-bit model, and both 4 and 8-bit models exhibit almost identical valid-response rates and overall performance.

\begin{figure}[ht!]
\begin{center}
\vskip 0.2in
\centerline{\includegraphics[width=\linewidth]{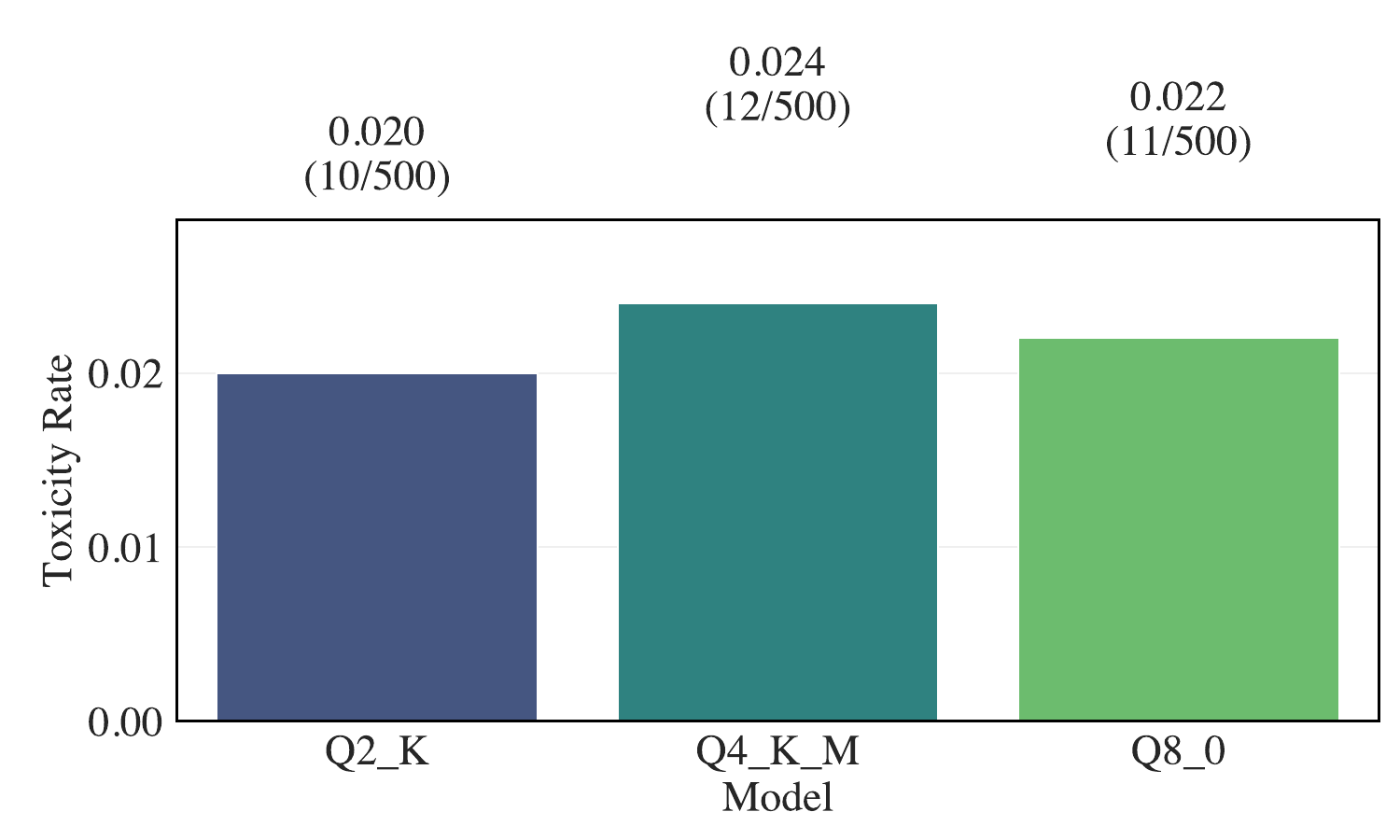}}
\caption{Toxicity rate by quantization level, measured with a DistilBERT-Base multilingual cased classifier}
\label{fig:toxicity_score}
\end{center}
\vskip -0.2in
\end{figure}

\paragraph{Toxicity:} Figure~\ref{fig:toxicity_score} reports the fraction of responses classified as toxic by a DistilBERT–Base multilingual cased toxicity classifier, over 500 toxicity‐prompt trials for each quantization level. The 2-bit model (Q2\_K) emits toxic content 2.0\% of the time (10/500), the 4-bit model (Q4\_K\_M) 2.4\% (12/500), and the 8-bit model (Q8\_0) 2.2\% (11/500). The 0.4 pp difference between the lowest and highest rates might indicate that aggressive quantization has minimal effect on the model's propensity to generate toxic language.

\end{document}